\documentclass[runningheads]{llncs}
\usepackage[T1]{fontenc}
\usepackage{graphicx}
\usepackage{booktabs}
\usepackage[misc]{ifsym}
\newcommand{\corr}{(\Letter)}
\usepackage{mwe}
\usepackage{amssymb}
\usepackage{xcolor,graphicx}

\begin{document}

\title{Disentangled VAD Representations via a Variational Framework for Political Stance Detection}

\titlerunning{PoliStance-VAE}

\author{Beiyu Xu\inst{1} \corr \and Zhiwei Liu\inst{1} \and
 Sophia Ananiadou\inst{1}}

\authorrunning{B Xu et al.}

\institute{The University of Manchester, Manchester, United Kingdom
\email{\{beiyu.xu,zhiwei.liu-2\}@postgrad.manchester.ac.uk}
\email{sophia.ananiadou@manchester.ac.uk}
}

\maketitle              

\begin{abstract}
The stance detection task aims to categorise the stance regarding specified targets. Current methods face challenges in effectively integrating sentiment information for stance detection. Moreover, the role of highly granular sentiment labelling in stance detection has been largely overlooked. This study presents a novel stance detection framework utilizing a variational autoencoder (VAE) to disentangle latent emotional features—value, arousal, and dominance (VAD)—from political discourse on social media. This approach addresses limitations in current methods, particularly in in-target and cross-target stance detection scenarios.  This research uses an advanced emotional annotation tool to annotate seven-class sentiment labels for P-STANCE. Evaluations on benchmark datasets, including P-STANCE and SemEval-2016, reveal that PoliStance-VAE achieves state-of-the-art performance, surpassing models like BERT, BERTweet, and GPT-4o. PoliStance-VAE offers a robust and interpretable solution for stance detection, demonstrating the effectiveness of integrating nuanced emotional representations. This framework paves the way for advancements in natural language processing tasks, particularly those requiring detailed emotional understanding.

\end{abstract}

\section{Introduction}
\begin{figure}[ht]
\centering
\includegraphics[width=\textwidth]{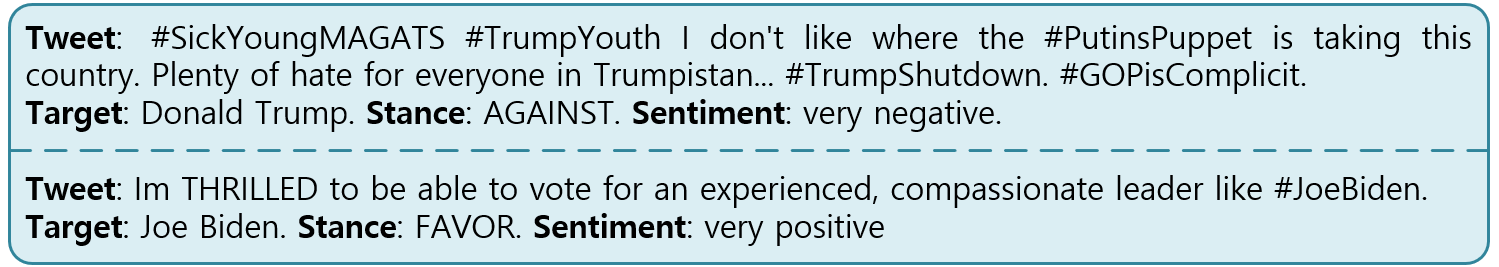}
\caption{Example of Data} 
\label{Data_example}
\end{figure}

In the era of pervasive social media influence, the analysis of online discourse has become a crucial area of study, particularly in the political domain \cite{alturayeif2023systematic}. Stance refers to the position, perspective, or attitude that a speaker or writer takes toward a particular topic \cite{kockelman2004stance}. It often reflects agreement, disagreement, support, opposition, or neutrality concerning a given claim or subject. Stance detection is widely used in misinformation detection, political opinion mining, fake news verification, and argumentation mining \cite{alturayeif2023systematic}. Political discussions on social media platforms like X, Snapchat, WeChat, and Reddit are often characterized by brevity and emotional undertones \cite{jaidka2019brevity,duncombe2019politics}. In Figure \ref{Data_example}, the present data indicates that the sentence structure is simple, with a pronounced sentiment obvious in the context. However, current methods that focus solely on context typically yield suboptimal results in stance detection.

Political comments on social media are often laden with sentiment, which can serve as an implicit indicator of stance as shown in Figure \ref{Data_example}. Previous studies \cite{sun2019stance,aldayel2019assessing,li2019multi} have explored the correlation between sentiment and stance using supervised learning approaches and pre-trained language models. However, these methods often fail to capture the nuanced interplay between sentiment and stance, particularly in multi-target or cross-target scenarios. Multi-target stance detection involves training a model on datasets containing multiple targets. Cross-target stance detection involves training and testing the model on different targets. Additionally, current datasets \cite{mohammad2016semeval,li2021p,sobhani2017dataset} for stance detection typically lack or possess low-granularity sentiment labels. Such situations further constrain the research of stance detection with sentiment information. 

To address the challenges mentioned above, this study proposes a novel variational autoencoder (VAE)-based framework, PoliStance-VAE, which incorporates sentiment classification as an auxiliary task. Furthermore, three latent emotional features—valence, arousal, and dominance (VAD)—are extracted from the text representation to capture the nuanced sentiment information. To facilitate the incorporation of nuanced sentiment information into the proposed model, an efficient large language model-based emotional annotation tool \cite{liu2024emollms} is employed to assign seven-class sentiment labels to the selected dataset. Evaluations on benchmark datasets, including P-STANCE \cite{li2021p} and SemEval-2016 \cite{mohammad2016semeval}, reveal that PoliStance-VAE achieves state-of-the-art performance, surpassing models like BERT, BERTweet, and GPT-4o.

The primary contributions of this work include: (1) the development of a VAE-based architecture for disentangling VAD features in political stance detection, (2) a comprehensive evaluation of the proposed model on P-STANCE and SemEval-2016 datasets, and (3) an ablation study highlighting the efficacy of fine-grained sentiment annotations in enhancing model performance. This approach not only addresses limitations in current stance detection methodologies but also paves the way for future research in integrating nuanced sentiment representations into NLP tasks.

\section{Related Work}
\subsection{Stance Detection}
Stance detection is a subtask within opinion mining that automatically classifies text stances concerning a specified target. Early studies concentrated on political debates and online forums, while contemporary research has shifted towards online social platforms like Twitter and Weibo. Classifying stances from social media data presents a challenge \cite{zhang2020enhancing}, primarily due to the diverse and informal characteristics like 
brevity, Abbreviations, and emoji. Recent research has attempted to detect the stance of discourses on social media. SemEval-2016 \cite{mohammad2016semeval} introduces a shared task focused on the detection of stance in tweets. The dataset in SemEval-2016 is widely utilised for subsequent research. Siddiqua et al. \cite{siddiqua2018stance} proposed a novel method for detecting stance in microblog posts by leveraging syntactic tree structures. The authors argue that syntactic tree representations effectively capture contextual and structural information in text, which enhances the performance of stance detection models. Chang Xu et al. \cite{xu2018cross} investigate the generalisation of classifiers across various targets through a neural model that utilises acquired knowledge from a source target to inform a destination target. Ji et al. \cite{ji2022cross} propose a many-to-one CTSD model utilising meta-learning techniques. They also enhance meta-learning by implementing a balanced and easy-to-hard learning pattern. The Stance Reasoner framework leverages language models with explicit reasoning paths to perform zero-shot stance detection on social media \cite{taranukhin2024stance}. The approach enhances interpretability by generating reasoning explanations for stance predictions and achieves competitive results across multiple datasets without task-specific fine-tuning.

\subsection{Stance Detection With Sentiment Information}
Affective information, including sentiments and emotions, is crucial. Such information help capture underlying feelings that influence stances, improving classification accuracy \cite{luo2022exploiting,zhang2020enhancing}. The study by Mohammad et al. \cite{sobhani2016detecting} explores methods for detecting stance in tweets and examines how stance interacts with sentiment in online discourse. The authors highlight that while stance and sentiment are related, they are distinct, and understanding both can provide deeper insights into opinions expressed on social media platforms. Smith and Lee \cite{sun2016exploring} investigate linguistic features pertinent to stance detection, focussing on the contributions of lexical, syntactic, and pragmatic cues in discerning a user's stance within text. The research indicates that the integration of various linguistic features, including modality, sentiment, and discourse markers, enhances the precision of stance classification across a range of topics and contexts. Sun et al. \cite{sun2019stance} introduced a joint neural network model designed to simultaneously predict the stance and sentiment of a post. The model first learns a deep shared representation between stance and sentiment information, and then sentimental information is utilized for stance detection through a neural stacking model. Huang and Yang \cite{huang2024multi} proposed a multi-stance detection model incorporating entiment features to improve the accuracy of stance classification. The model uses an LDA-based five-category stance indicator system, extracts sentiment features via a lexicon, and applies a hybrid neural network for precise stance detection. This study utilises sentiment as the supervisory signal for disentangled representation learning, distinguishing it from prior research, owing to the architecture of the variational autoencoder. An auxiliary task for sentiment prediction is incorporated into the proposed model. PoliStance-VAE effectively integrates sentiment information with contextual information for stance detection through disentangled variables.

\section{METHODOLOGY}
\subsection{Task Definition}
The primary objective is identifying the stance labels associated with the provided tweets. The auxiliary task is to categorise the sentiment expressed in the provided tweets. Dataset $D$ comprises short text $\{t_{1},t_{2}, ..., t_{n}\}$, each associated with corresponding ground-truth stance labels $y_{i}$ and sentiment label $e_{i}$, where $y_{i}\in S$ and $e_{i}\in E$, with $S$ and $E$ representing the pre-defined stance and sentiment label sets, respectively. Each $t_{i}$ comprises $m_{i}$ tokens: $\{ t^{1}_{i},t^{2}_{i}, ..., t^{m}_{i}\}$. Additionally, each $t_{i}$ connects with a target $g_{i}$, where $g_{i}$ belongs to the pre-defined target set $G$. With the above information, the tasks can be formalised as $\hat{y}_{i}, \hat{e}_{i} = f(t_{i},g_{i})$, where $f$ refers to trainable parameters of the model.

\begin{figure}[t]
\includegraphics[width=\textwidth]{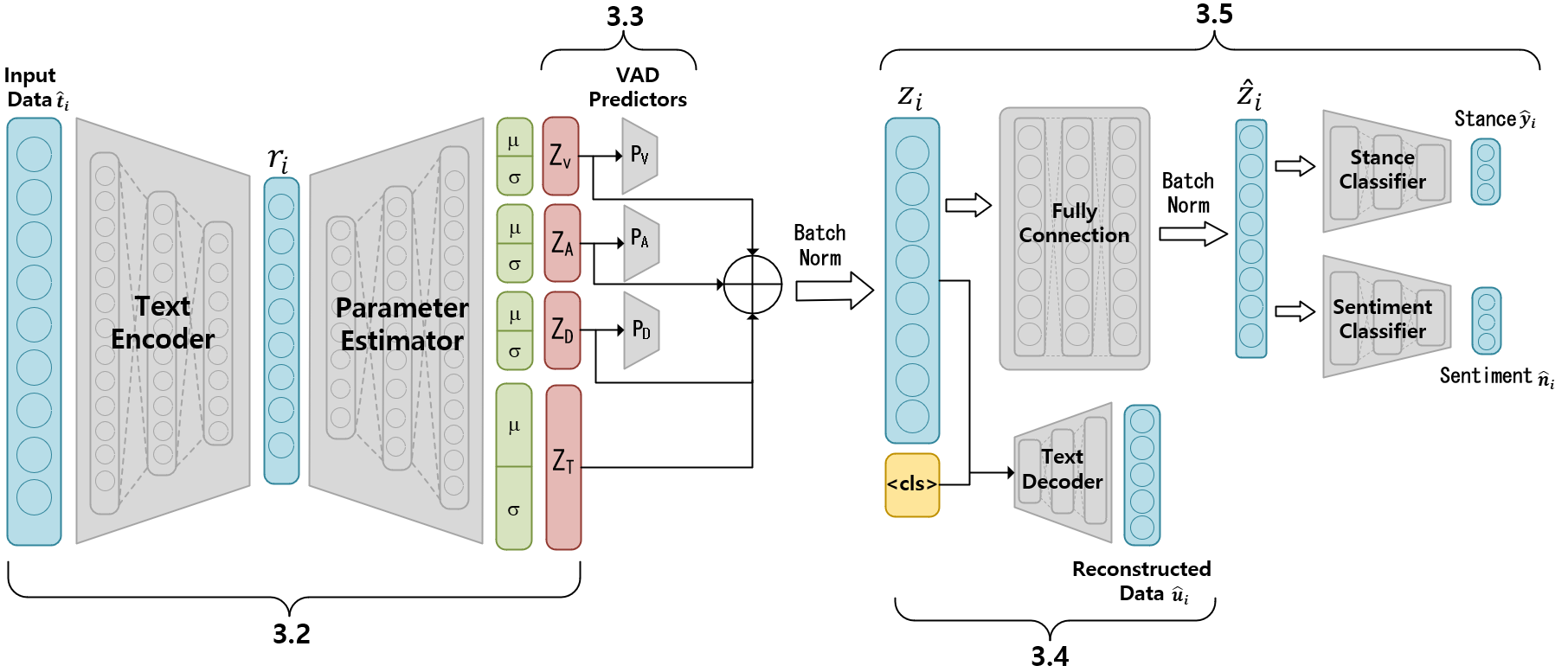}
\caption{Main components of PoliStance-VAE.} 
\label{backbone}
\end{figure}

\subsection{Disentangled Representations}
This section outlines the process of disentangling latent variables. Using the text encoder, we extract four latent representations from the text representations, which form the basis of PoliStance-VAE, as illustrated in Figure \ref{backbone}.

\subsubsection{Target-Aware Text Encoder}
To explicitly introduce target information, we position the target $g_{i}$ at the end of the text to obtain the target-aware input $\hat{t}_{i}$: 
\begin{equation}
\hat{t}_{i} = \{\langle cls \rangle; t^{1}_{i}; t^{2}_{i}; ...; t^{m}_{i}; \langle sep \rangle ; g_{i}; \langle eos \rangle \}
\end{equation}
where \{;\} indicates concatenation, $\langle cls \rangle$ and $\langle sep \rangle$ represent the start-of-sentence and end-of-sentence tokens, respectively. The $\langle sep \rangle$ token is inserted between the target and the text to separate these two components. By employing a PLM-based encoder, target-aware text embeddings are obtained. 
\begin{equation}
r_{i} = Encoder(\hat{y}_{i})
\end{equation}
where $Encoder$ refers to the RoBERTa-Large \cite{liu2019roberta} text encoder, $r_{i}\in \mathbb{R}^{L\times D_{h}}$ represents the text representation, $L$ indicates the sequence length, and $D_{h}$ denotes the dimension of the hidden states. This study utilises the embedding of the start-of-sentence token: $r^{[cls]}_{i} \in \mathbb{R}^{D_{h}}$, as the text-level representation of $t_{i}$.

\subsubsection{Parameter Estimator}
The study of disentangled variational autoencoders for emotion recognition in conversations informs the design of the generative model architecture \cite{yang2023disentangled}. Three latent features, Valence, Arousal, and Dominance (VAD), are extracted from the hidden text representations. Valence refers to the pleasantness of a stimulus, Arousal denotes the intensity of the emotional response elicited by a stimulus, and Dominance indicates the degree of control a stimulus exerts \cite{warriner2013norms}. Additionally, a "Content" feature: $T$ is defined to regulate the content generation of the target text.

The initial VAE model presents a practical estimator for the lower bound and its derivatives concerning the parameters \cite{kingma2013auto}. The deterministic text representation is substituted with an approximation of the posterior $q_{\phi}(z|x)$, parameterised by a neural network. In this study, Four feed-forward neural networks are utilised to map $x = r^{[cls]}_{i}$ to four sets of parameters for Gaussian distributions $(\mu,\sigma)$. The latent representation of each feature can be sampled from the Gaussian distribution characterised by the corresponding $(\mu_{(R)},\sigma_{(R)})$ through the reparametrization trick \cite{warriner2013norms}:
\begin{equation}
Z_{(i,R)} = \mu_{(i,R)} + \sigma_{(i,R)} \odot \epsilon
\end{equation}
where $\epsilon$ is an auxiliary noise variable $\epsilon \sim N(0,1)$, $R \in \{ V, A, D, T\}$ is the latent parameter set. Then, the latent representations are concatenated: $Z_{i} = [Z_{V};Z_{A};Z_{D};Z_{T}]$.

\subsection{Supervised Learning of VAD Disentangled Representation}
Our objective is to improve the disentangled VAD representations, ensuring they contain sufficient information to accurately predict the associated generative factor. Thus, we introduce supervised learning to enhance the informativeness of VAD latent representations. 

To augment the representation's capacity to predict the associated generative factor, we incorporate supervisory signals derived from the NRC-VAD, a sentiment lexicon that provides dependable human evaluations of Valence, Arousal, and Dominance (VAD) for 20,000 English terms. All terms within the NRC-VAD framework signify or imply emotions and are derived from widely utilised sentiment lexicons and tweets. An aggregation process evaluates each term for VAD on a scale ranging from 0 to 1. In practical applications, such signals can efficiently convert sentiment labels into their corresponding numerical ratings. For stance, the sentiment $positive = \{V:0.959,A:0.510,D:0.855\}$. Since fine-grained VAD supervision signals are introduced, we can obtain the VAD score for each sentiment term $e_{i}$ within the pre-defined categorical sentiment set $E$. 
\begin{equation}
vad_{e_{i}} = \{vad^{V}_{e_{i}},vad^{A}_{e_{i}},vad^{D}_{e_{i}} \}
\end{equation}
The enhancement of VAD representations is anticipated to improve model performance in stance detection.

The VAD prediction is specifically derived from the latent representation $Z_{(i, I)}$ through the application of a feed-forward neural network. A sigmoid activation function is employed to transform the prediction into the range of (0, 1):
\begin{equation}
P_{(i,I)} = \frac{1}{1+e^{-( Z_{(i,I)} W_I+b_I) }}
\end{equation}
where $W_I$ and $b_I$ are parameters of the predictor, and $I \in \{V,A,D\}$. For the training objective, the mean squared error loss is calculated between the predictions and signals:
\begin{equation}
\mathcal{L}_{VAD}(\phi,\kappa) = \frac{1}{N} \sum^{N}_{i=1} \sum_{I} (P_{(i,I)} - vad^{I}_{e_i})
\end{equation}
where $\phi$ and $\kappa$ denote the parameters of the encoder and the VAD predictor, $e_{i}$ refers to the sentiment label, and N is batch size.

\subsection{Text Reconstruction}
To reconstruct the input data, a batch normalisation is applied to $Z_{i}$ after concatenation. The application of batch normalisation to the concatenated latent representations contributes to improved training efficiency, stability, and generalisation. Then, we feed the $Z_{i}$ and target $t_{i}$ into the $Decoder$:
\begin{equation}
\hat{u}_{i} = Decoder(Z_{i},t_{i})
\end{equation}
where $Decoder$ is the BART-Large model \cite{lewis2019bart}, and $\langle eos \rangle$ is taken out of $t_{i}$. In a standard VAE, a KL-divergence term is incorporated for each latent space to maintain proximity between the approximate posterior and the prior distribution. During training, the Evidence Lower Bound (ELBO) is leveraged as the training objective \cite{yang2023disentangled}:
\begin{equation}
\mathcal{L}_{ELBO}(\phi,\theta;x_{i}) = -\mathbb{E}_{q_{\phi}(z_{i}|x_{i})}[log\,p_{\theta}(x_{i}|z_{i})] + \sum_{R} \delta KL[q^{R}_{\phi}(z_{(i,R)}|x_{i})\,||\, p(z_{(i,R)})]
\end{equation}
where $\phi$ and $\theta$ refer to the parameters of the encoder and decoder, $\delta$ is a coefficient to control the KL-divergence term, and $p(z_{(i,R)})$ is calculated with standard Gaussian prior.

\subsection{Model Training Objective}
The final latent representation $\hat{Z}_{i}$ is derived by inputting $Z_{i}$ into a fully connected layer followed by a batch normalisation layer:
\begin{equation}
\hat{Z}_{i} = Batch\,Norm(Z_{i}W_{f}+b_{f})
\end{equation}
where $W_f$ and $b_f$ denote the parameters of the fully connected layer. Subsequently, the latent representation $\hat{Z}_i$ is utilised to compute the classification probability of the stance label and sentiment label:
\begin{equation}
\hat{y}_i = Softmax(\hat{Z}_{i}W_s+b_s)
\end{equation}
\begin{equation}
\hat{n}_i = Softmax(\hat{Z}_{i}W_j+b_j)
\end{equation}
where $W_s$, $b_s$, $W_j$, and $b_j$ are learnable parameters. Then, the stance detection and sentiment classification losses are computed using standard cross-entropy loss:
\begin{equation}
\mathcal{L}_{sd} = -\frac{1}{N}\sum^{N}_{i=1}\sum^{|S|}_{j=1}y^{j}_{i}log\,(\hat{y}^{j}_{i})
\end{equation}
where $y^{j}_{i}$ and $\hat{y}^{j}_{i}$ are j-th element of $y_{i}$ and $\hat{y}_{i}$.
\begin{equation}
\mathcal{L}_{sent} = -\frac{1}{N}\sum^{N}_{i=1}\sum^{|E|}_{j=1}n^{j}_{i}log\,(\hat{n}^{j}_{i})
\end{equation}
where $n^{j}_{i}$ and $\hat{n}^{j}_{i}$ are j-th element of $n_{i}$ and $\hat{n}_{i}$. Finally, we combine all modules and formulate a multi-task loss:
\begin{equation}
\mathcal{L} = \mathcal{L}_{sd} + \alpha_{sent}\mathcal{L}_{sent} + \alpha_{ELBO}\mathcal{L}_{ELBO} + \alpha_{VAD}\mathcal{L}_{VAD}
\end{equation}
where the $\alpha$s are the pre-defined weight coefficients.

\section{Experiments}
Stance detection can be categorised based on the number and presence of targets in datasets into three types: target-specific (in-target) stance detection \cite{mohammad2016semeval}, multi-target stance detection \cite{sobhani2017dataset}, and cross-target stance detection \cite{li2021p}. This study employs in-target and cross-target stance detection on two benchmark datasets to assess the model's performance. 

\subsection{Datasets}
\subsubsection{P-STANCE}
The P-STANCE \cite{li2021p} dataset is a large annotated dataset designed for stance detection in the political domain. It consists of 21,574 tweets collected during the 2020 U.S. presidential election, focusing on three political figures: Donald Trump, Joe Biden, and Bernie Sanders. Each tweet is annotated with a stance label, either 'Favor' or 'Against'. Table \ref{dataset:P-STANCE} demonstrates the Label distribution across different targets for P-STANCE. 
\begin{table}[ht]
\centering
\caption{Data distribution across different targets in P-STANCE.}
\begin{tabular}{l@{\hskip 15pt}l|r@{\hskip 15pt}r@{\hskip 15pt}r}
\hline
    &      & \textbf{Trump} & \textbf{Biden} & \textbf{Sanders} \\ 
\hline
\textbf{Train}   
    & Favor   & 2,937 & 2,552 & 2,858 \\ 
    & Against & 3,425 & 3,254 & 2,198 \\ 
\textbf{Val}    
    & Favor   & 365   & 328   & 350   \\ 
    & Against & 430   & 417   & 284   \\ 
\textbf{Test}    
    & Favor   & 361   & 337   & 343   \\ 
    & Against & 435   & 408   & 292   \\ 
\hline
\textbf{Total}  
    &         & 7,953 & 7,296 & 6,325 \\ 
\hline
\end{tabular}
\label{dataset:P-STANCE}
\end{table}

\noindent Furthermore, the dataset was annotated with seven sentiment labels utilising Emollama-chat-13b within the EmoLLMs framework. EmoLLMs \cite{liu2024emollms} represent a collection of advanced emotional large language models and annotation tools designed for in-depth affective analysis. These tools demonstrate superior performance compared to all other open-sourced LLMs and exceed ChatGPT and GPT-4 in the majority of tasks on extensive affective evaluation benchmarks.

\subsubsection{SemEval-2016}
SemEval-2016 \cite{mohammad2016semeval} comprises six targets: Atheism, Climate Change as a Real Concern, Feminist Movement, Hillary Clinton, Legalization of Abortion, and Donald Trump. The dataset is labelled for the purpose of stance detection regarding a specified target. The data associated with Donald Trump is excluded, as it is designated for a different stance detection task. In this study, the original training set is divided into the new training and validation sets for hyperparameter tuning. This dataset also annotates the data with three sentiment labels. Table \ref{dataset:SemEval} illustrates the data distribution of the new version.
\begin{table}[ht]
\centering
\caption{Data distribution across different targets in SemEval-2016.}
\begin{tabular}{l @{\hskip 15pt} l|r @{\hskip 15pt}r @{\hskip 15pt}r @{\hskip 15pt}r @{\hskip 15pt}r} 
\hline
    &      & \textbf{Atheism} & \textbf{Climate} & \textbf{Feminist} & \textbf{Hillary} & \textbf{Abortion} \\ 
\hline
\textbf{Train}   
    & Favor   & 83  & 195  & 190  & 110  & 109  \\ 
    & Against & 280 & 14   & 286  & 341  & 326  \\ 
    & Neither & 108 & 150  & 119  & 155  & 157  \\ 
\textbf{Val}     
    & Favor   & 9   & 17   & 20   & 8    & 12   \\ 
    & Against & 24  & 1    & 42   & 52   & 29   \\ 
    & Neither & 9   & 18   & 7    & 23   & 20   \\ 
\textbf{Test}    
    & Favor   & 32  & 123  & 58   & 45   & 46   \\ 
    & Against & 160 & 11   & 183  & 172  & 189  \\ 
    & Neither & 28  & 35   & 44   & 78   & 45   \\ 
\hline
\textbf{Total}   
    &         & 733 & 564  & 949  & 984  & 933  \\ 
\hline
\end{tabular}
\label{dataset:SemEval}
\end{table}

\subsection{Sentiment Analysis}
This part illustrates the correlation between stance and sentiment through the analysis of sentiment distribution across targets. 

In the P-STANCE dataset, the "AGAINST" stance primarily demonstrates negative sentiments, including "moderate negative" and "very negative," indicating critical or dissenting attitudes towards the target. The "FAVOR" stance is characterised by positive sentiments, including "moderate positive" and "slightly positive," reflecting supportive or favourable opinions regarding the target. Neutral or balanced sentiments, such as "slightly positive" or "moderate negative," are present in both stance labels but in differing proportions, indicating nuanced expressions even within the distinct stance. 
\begin{figure}[ht]
    \centering
    \includegraphics[width=0.7\textwidth]{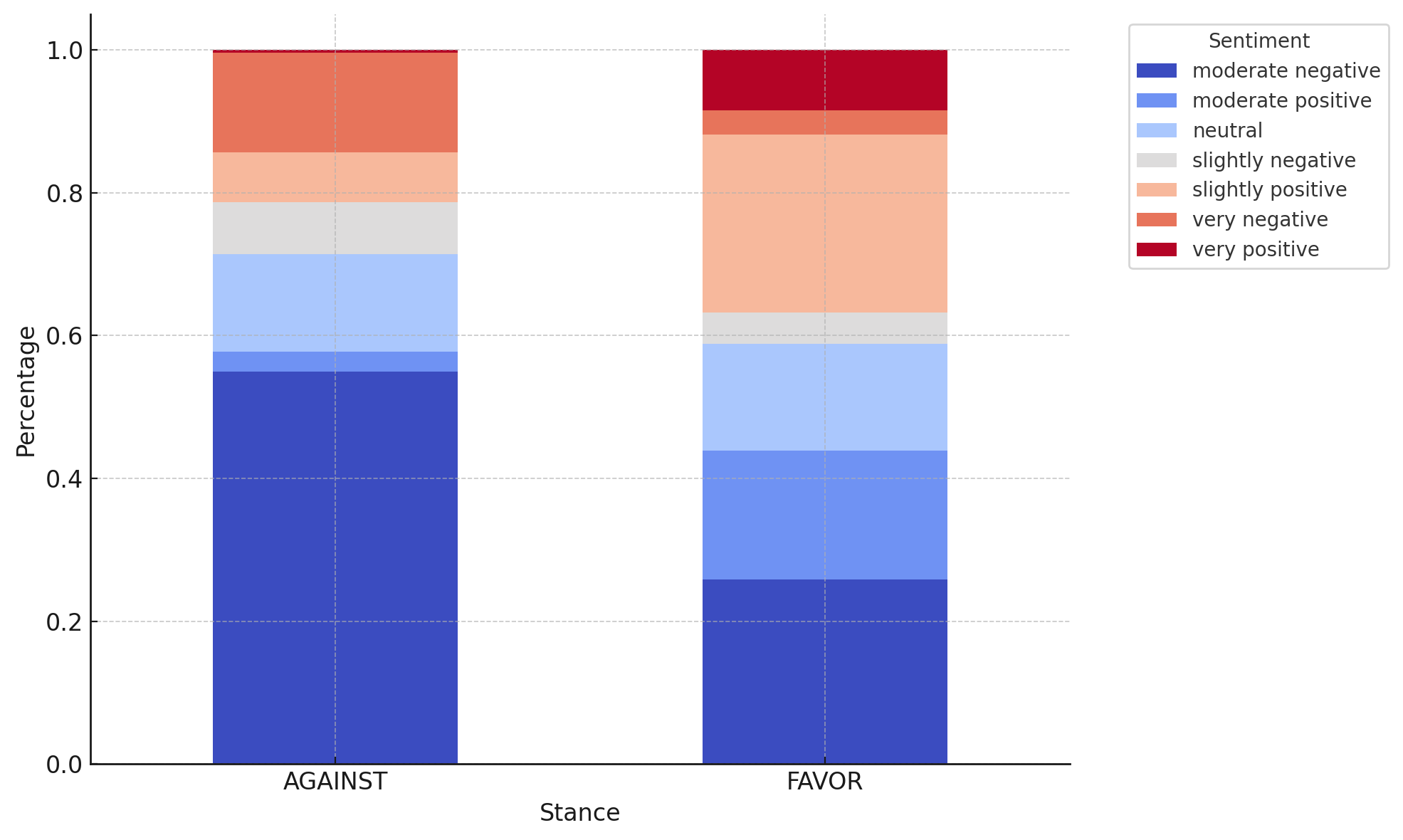}
    \caption{Sentiment distribution across the targets in P-STANCE.}
    \label{Sentiment_PStance}
\end{figure}

\noindent For the SemEval-2016 dataset, the "AGAINST" stance is primarily associated with negative sentiment, reflecting a critical or oppositional tone, while positive and neutral sentiments are less common. The "FAVOR" stance exhibits a mix of negative and positive sentiments, suggesting a blend of support and critique, with neutrality appearing occasionally. The "NONE" stance demonstrates a more balanced distribution of sentiments, showing the highest occurrence of neutrality compared to other stances.

\begin{figure}[ht]
    \centering
    \includegraphics[width=0.7\textwidth]{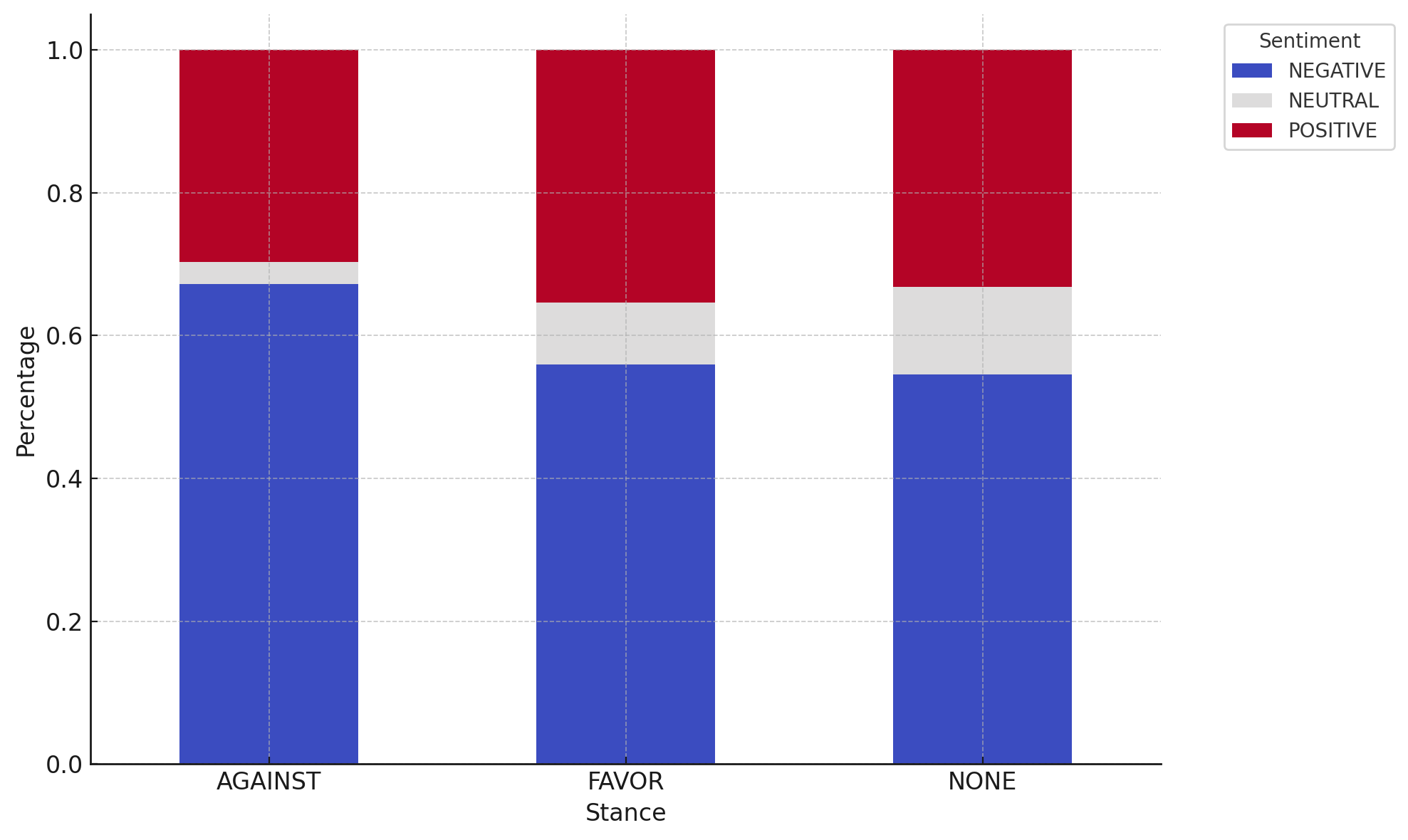}
    \caption{Sentiment distribution across the targets in SemEval-2016.}
    \label{Sentiment_SemEval}
\end{figure}

In summary, there is a strong correlation between sentiment and stance; supportive stances are associated with positive sentiments, whereas opposing stances are inclined towards negative sentiments. This pattern indicates that stance labels frequently embody a fundamental sentiment direction, which can be utilised for stance detection.

\subsection{Evaluation Metrics}
The evaluation metrics utilised are identical to those presented in the P-STANCE and SemEval-2016 papers. The macro-average of the F1-score for 'favour' and the F1-score for 'against' served as the primary evaluation metric.
\begin{equation}
F_{avg} = \frac{F_{favor}+F_{against}}{2}
\end{equation}
where $F_{favor}$ and $F_{against}$ are computed as follows:
\begin{equation}
F_{favor} = \frac{2P_{favor}R_{favor}}{P_{favor}+R_{favor}}
\end{equation}
\begin{equation}
F_{against} = \frac{2P_{against}R_{against}}{P_{against}+R_{against}}
\end{equation}

\subsection{Baseline}
This section presents a detailed comparison of our approach with various baseline models. The baselines are categorised into pre-trained language models (PLMs), traditional models (TMs), and large language models (LLMs).
\begin{itemize}
    \item \textbf{TMs}: BiLSTM \cite{schuster1997bidirectional}.\\
    The BiLSTM framework which takes tweets as inputs without considering the target information. The tweets are encoded using bert-based embeddings. 
    \item \textbf{PLMs}: Bert \cite{kenton2019bert}, BERTweet \cite{nguyen2020bertweet}. \\
    BERT and BERTweet are fine-tuned to predict the stance by appending a linear classification layer to the latent representation of the [CLS] token.
    \item \textbf{LLMs}: GPT-4o \cite{openai2024gpt4}, GPT-3.5-turbo \cite{openai2023chatgpt}. \\
    GPT series are conducted for zero-shot stance detection using the following prompts.
\end{itemize}

\begin{center}
\fcolorbox{black}{gray!10}{
\begin{minipage}{0.9\textwidth}
\footnotesize
\textbf{Template}  \\
\textbf{System Prompt}: You are tasked with classifying the stance of a given text (Tweet) concerning a specific target. The stance can be one of the following: \\
\textit{FAVOR: The text supports the target.}  \\
\textit{AGAINST: The text opposes the target.}  \\
\textit{NONE: The tweet is neutral or doesn't have a stance towards the target.}
Consider the context of the text and classify its stance.
\textbf{User prompt}:
\newline
Target: [Target]. Text: [Text]. What is the stance? Please reply with only one word: FAVOR, AGAINST or NONE.
\end{minipage}
}
\end{center}

\subsection{Implementation Details}
All experiments are conducted utilising a single Nvidia Tesla A100 GPU, which has 80GB of memory. We initialise the pre-trained weights of all PLMs and utilise the tokenisation tools provided by Hugging Face \cite{wolf2019huggingface}. AdamW optimiser \cite{loshchilov2017decoupled} is used to train the model. All hyper-parameters are tuned on the validation set. For the baseline models, the learning rate is $2e^{-5}$. Batch size is 32. For Polistance-VAE, the learning rate of P-STANCE is $8e^{-6}$, and the batch size is 32. The learning rate of SemEval-2016 is $1e^{-5}$, and the batch size is 16.

\section{RESULTS AND ANALYSIS}
This section presents the experiments conducted on various stance detection tasks using the dataset, along with the results obtained from the specified baselines. The results presented are the averages derived from five runs utilising distinct random seeds. Additionally, we compare the results with relevant studies. 

\subsection{In-Target Stance Detection}
In-target stance detection refers to a stance detection task in which a classifier is both trained and validated on the same target. Most prior studies utilise an "Ad-hoc" training strategy, training a distinct model for each target and evaluating it on the corresponding test set. However, the model tends to predict the stance by adhering to specific patterns, often neglecting the target information and resulting in overfitting. This study also conducts experiments utilising the "Merged" training strategy proposed in \cite{li2021p}, wherein a model is trained and validated on all targets while tested on separate targets. The SemEval-2016 dataset exhibits limited and unbalanced data for each target. We exclusively perform the "Merged" experiments for this purpose. 

\begin{table}[!ht]
\centering
\caption{Results of Ad-hoc and Merged experiments on P-STANCE. Best values are highlighted in bold.}
\begin{tabular}{l@{\hskip 15pt}|r@{\hskip 15pt}r@{\hskip 15pt}r@{\hskip 15pt}r}
\hline
  \textbf{Method/Ad-hoc}  & \textbf{Trump} & \textbf{Biden} & \textbf{Sanders} & \textbf{F$_{avg}$}\\ 
\hline

\textbf{BiLSTM} \cite{li2021p}  & 76.92   & 77.95 & 69.75 & 74.87 \\

\textbf{BERT} \cite{li2021p}  & 78.28   & 78.70 & 72.45 & 76.48 \\

\textbf{BERTweet} \cite{li2021p}  & 82.48   & 81.02 & 78.09 & 80.53 \\

\textbf{ChatGPT} \cite{zhang2022would}    & 83.20   & 82.00 & \textbf{79.40} & 81.50 \\ 

\textbf{PoliStance-VAE} & \textbf{86.29}   & \textbf{85.69}   & 79.39 & \textbf{83.79}   \\ 

\hline

\textbf{Method/Merged} &    &   &   & \\
\hline

\textbf{BiLSTM} \cite{li2021p}  & 77.18   & 75.47 & 67.43 & 73.36 \\

\textbf{BERT} \cite{li2021p}  & 79.19   & 76.02 & 73.59 & 76.27 \\

\textbf{BERTweet} \cite{li2021p}  & 83.81  & 79.08 & 77.75 & 80.21 \\ 

\textbf{CKD} \cite{li2023distilling}   & -  & - & - & 81.97 \\ 

\textbf{GPT-3.5}   & 76.05  & 82.91 & 79.51 & 78.74 \\ 

\textbf{GPT-4o}   & 83.82  & 82.83 & \textbf{79.73} & 81.79 \\ 

\textbf{PoliStance-VAE} & \textbf{85.37}   & \textbf{84.23}  & 78.12  & \textbf{82.57}   \\ 
\hline
\end{tabular}
\label{results PStance}
\end{table}

Experimental results of these two different settings are shown in Table \ref{results PStance}. The experimental results demonstrate that the PoliStance-VAE model consistently outperforms other approaches in both "Ad-hoc" and "Merged" training strategies, achieving the highest average F-scores of 83.79 and 82.57, respectively.
However, ChatGPT and GPT-4o obtained the best performance on Bernie Sanders, which indicates the potential of the large language models. Performance drops can be observed in the "Merged" setting, which indicates such a strategy can test whether the model learns target-specific representations. Moreover, PoliStance-VAE has the minimum performance drop compared to baseline models, which demonstrates the proposed method can better capture the target-specific representations.

\begin{table}[!ht]
\centering
\caption{Expermients using "Merged" training strategy for SemEval-2016. AT: Atheism, FM: Feminist Movement, HC: Hillary Clinton, LA: Legalization of Abortion, CC: Climate Change as a Real Concern. Best values are highlighted in bold.}
\begin{tabular}{l@{\hskip 15pt}|r@{\hskip 15pt} r @{\hskip 15pt}r @{\hskip 15pt}r @{\hskip 15pt}r @{\hskip 15pt}r} 
\hline
\textbf{Method}       & \textbf{AT} & \textbf{FM} & \textbf{HC} & \textbf{LA} & \textbf{CC} & \textbf{F$_{avg}$}\\ 
\hline
\textbf{BiLSTM}  
    & 42.11   & 39.10  & 36.83  & 40.30  & 6.11  & 36.41  \\
\textbf{BERT} 
    & 59.24   & 55.00  & 55.88  & 64.04  & 45.77  & 68.45  \\
\textbf{BERTweet} 
    & 56.40   & 53.56  & 57.49  & 58.44  & 45.35  & 67.89  \\
\textbf{GPT-3.5} 
    & 14.59   & 68.65  & 69.97  & 57.19  & 62.45  & 58.84  \\
\textbf{GPT-4o}  
    & 33.68   & \textbf{72.42}  & 66.79  & \textbf{68.79}  & 64.66  & 71.59  \\ 
\textbf{TakeLab} \cite{wei2016pkudblab}  
    & 66.83 & 67.25 & 41.25   & 53.01  & \textbf{67.12}  & 61.38  \\ 
\textbf{DeepStance} \cite{vijayaraghavan2016deepstance} 
    & 63.54 & 52.90 & 40.40  & 52.34  & 55.35  & 63.32 \\ 
\textbf{IDI@NTNU}  \cite{bohler2016idi}
    & 62.47 & 59.59 & 54.86  & 48.59  & 57.89  & 54.47  \\ 
\textbf{PoliStance-VAE}   
    & \textbf{73.49} & 64.61 & \textbf{73.31}  & 64.17 & 54.96  & \textbf{74.01}   \\ 
\hline
\end{tabular}
\label{results SemEval}
\end{table}

\noindent Table \ref{results SemEval} presents the results of the experiment conducted on SemEval-2016. The PoliStance-VAE model achieved an average F1-score of 74.01 across five targets. This performance exceeds multiple baseline models, including BiLSTM, BERT, and BERTweet, along with earlier methods such as TakeLab and DeepStance. GPT-3.5 attained an average F1-score of 58.84, whereas GPT-4o recorded a higher score of 71.59. Both models exhibited competitive performance, especially GPT-4o, which surpassed several baselines including BiLSTM (36.41), BERT (68.45), and BERTweet (67.89). Notably, PoliStance-VAE outperformed the highest-ranking system from the original SemEval-2016 Task 6 \cite{mohammad2016semeval}, which reported an F1-score of 67.82.

This comparison highlights the advantages of the PoliStance-VAE model's incorporation of disentangled Valence-Arousal-Dominance (VAD) representations and its robustness in capturing nuanced stance-related features, even when tested against advanced large language models like GPT-4o.

\subsection{Cross-Target Stance Detection}
Conventional models for stance detection exhibit limited generalisation capabilities when applied to data from new targets, hence resulting in research into cross-target stance detection. The model for cross-target stance detection is initially trained and validated on a source target before being tested on a destination target. This subsection presents the results of cross-target experiments conducted on the P-STANCE dataset. The paper \cite{li2021p} of P-STANCE provides the performance metrics of various models, serving as a baseline for this study. The results are demonstrated in Table \ref{results cross-target}.

\begin{table}[!ht]
\centering
\caption{Cross-Target experiments on P-STANCE. DT: Donald Trump, JB: Joe Biden, BS: Bernie Sanders. Best values are highlighted in bold.}
\begin{tabular}{l@{\hskip 10pt}|r@{\hskip 10pt}r@{\hskip 10pt}r@{\hskip 10pt}r}
\hline
  \textbf{Target}  & \textbf{BiCE} & \textbf{CrossNet} & \textbf{BERTweet} & \textbf{PoliStance-VAE}\\ 
\hline

\textbf{DT$\rightarrow$JB}  
& 55.83   & 56.67 & 58.88 & \textbf{61.10} \\

\textbf{DT$\rightarrow$BS} 
& 51.78   & 50.08 & 56.50 & \textbf{56.84} \\

\textbf{JB$\rightarrow$DT}  
& 58.16   & 60.43 & 63.64 & \textbf{68.86} \\

\textbf{JB$\rightarrow$BS} 
& 60.24   & 60.81 & 67.04 & \textbf{73.66} \\

\textbf{BS$\rightarrow$DT} 
& 51.41   & 52.99 & 58.75 & \textbf{62.27} \\

\textbf{BS$\rightarrow$JB} 
& 57.68   & 62.57 & 72.99 & \textbf{79.37} \\

\textbf{DT\&JB$\rightarrow$BS} 
& 52.26   & 56.26 & 69.99 & \textbf{73.76} \\

\textbf{DT\&BS$\rightarrow$JB} 
& 53.73   & 55.57 & 68.64 & \textbf{75.28} \\

\textbf{JB\&BS$\rightarrow$DT} 
& 53.91   & 56.44 & 66.01 & \textbf{66.53} \\
\hline
\end{tabular}
\label{results cross-target}
\end{table}

\noindent Traditional models, including BiCE and CrossNet, demonstrated restricted generalisation abilities. BERTweet outperforms others by acquiring more universal representations through its advanced architecture. PoliStance-VAE outperforms BERTweet across all cross-target tasks, indicating superior generalisation capability. In summary, PoliStance-VAE introduces disentangled Valence-Arousal-Dominance (VAD) representations to enhance cross-target stance detection by capturing hidden sentiment information embedded in the text. This innovative approach leverages advanced architecture to achieve superior generalization capabilities, outperforming strong baselines like BERTweet across all cross-target tasks. These results underscore the importance of integrating disentangled representation learning for improved performance in cross-target stance detection, offering a significant advancement over traditional models.

\subsection{Ablation Study}
This section presents an ablation analysis in Table \ref{results ablation} to examine the impact of each module. The training strategy is designated as "Merged" and in-target stance detection, as both datasets can be effectively compared under these conditions. The "Decoder" denotes the module responsible for reconstructing the input sentence. "VAD" denotes the module designed for supervised learning of VAD disentangled representation. "Sentiment" refers to the auxiliary task of classifying the sentiment expressed in the text.

\begin{table}[!ht]
\centering
\caption{Ablation analysis on both datasets. Best values are highlighted in bold. "w/o" means without.}
\begin{tabular}{l@{\hskip 15pt}|@{\hskip 15pt}c@{\hskip 15pt}c}
\hline
  \textbf{Method}  & \textbf{P-STANCE} & \textbf{SemEval-2016}\\ 
\hline

\textbf{PoliStance-VAE}  
& 82.57   & 74.01  \\

\textbf{w/o Decoder} 
& 83.20($\uparrow$0.63)   & 73.80($\downarrow$0.21)  \\

\textbf{w/o VAD} 
& 67.12($\downarrow$15.45)   & 69.02($\downarrow$4.99)  \\

\textbf{w/o Sentiment} 
& 77.14($\downarrow$5.43)   & 73.96($\downarrow$0.05)  \\

\hline
\end{tabular}
\label{results ablation}
\end{table}
\noindent The results indicate a significant performance decline when the VAD latent variables are removed, suggesting that these variables capture essential hidden and shared information across targets. The outcome of P-STANCE in the absence of the decoder shows a marginal increase. The performance of the same settings on SemEval-2016 exhibits a slight decline. The reason for this occurrence may be the brevity and simplicity of the sentence structure. The data from the two datasets were sourced from Twitter. Comments on social media typically exhibit a straightforward structure and articulate clear points. The model can acquire limited unique and concealed contextual information. The sentiment module contributes positively to stance detection, as reflected in improved performance metrics. The model's performance on SemEval-2016 exhibited a slight decline in the absence of sentiment classification. A significant decline is observed in both the settings of without VAD and Sentimen when the model is trained on P-STANCE. The tri-categorical sentiment labels could limit the model's ability to learn. Utilising more granular sentiment labels could improve the model's ability to capture nuanced sentiment information from text. Additionally, more nuanced sentiment labels produce a wider range of VAD scores for the text, which may facilitate the learning of disentangled variables. Therefore, these results show that the proposed model benefits more from fine-grained information.

\section{Conclusion}
This paper presents a method utilising a variational autoencoder that incorporates sentiment information for stance detection. We introduce an auxiliary sentiment classification task to assist the model in capturing hidden information. Subsequently, the three hidden features—Valence, Arousal, and Dominance—are disentangled from the latent space. The sentiment of the text serves as the label for supervised learning of latent features. The disentangled features, along with contextual information, are utilised to reconstruct the sentences.

Experimental results indicate that PoliStance-VAE surpasses the leading model on two benchmark datasets for political stance detection. The ablation study demonstrates the efficacy of the proposed auxiliary sentiment classification task and the supervised learning of disentangled variables. Furthermore, fine-grained sentiment annotations may enhance the performance of stance detection. Given the large amounts of data, an affective analysis tool annotates the sentiment labels of P-STANCE. Further work on sentiment annotation by humans may bring a more accurate result. In the future, we anticipate working with more fine-grained sentiment information from datasets, which may ultimately better capture the hidden and disentangled features.

\begin{credits}
\subsubsection{\ackname} 
This work is supported by the computational shared facility at the University of Manchester. 

\subsubsection{\discintname}
The authors have no competing interests to declare that are relevant to the content of this article.
\end{credits}
%
%
%
\bibliographystyle{splncs04}
\bibliography{main}
%

\end{document}